\documentclass[runningheads]{llncs}
\usepackage{graphicx}
\usepackage[numbers]{natbib}
\usepackage{amsmath,amssymb,amsfonts}
\usepackage{algorithmic}
\usepackage[ruled]{algorithm2e}
\usepackage{graphicx}
\usepackage{textcomp}
\usepackage{xcolor}
\usepackage[T1]{fontenc} 
\usepackage{amsmath}
\usepackage[cmintegrals]{newtxmath}
\usepackage{bm} 
\begin{document}
\title{ Automatic Relevance Determination Bayesian Neural Networks for Credit Card Default Modelling}
\titlerunning{Bayesian Neural Networks for Credit Card Default Modelling}
%
\author{Rendani Mbuvha \and
Illyes Boulkaibet\and
Tshilidzi Marwala}
\authorrunning{R. Mbuvha et al.}
%
\institute{ School of Electrical and Electronic Engineering\\
University of Johannesburg\\
Johannesburg, South Africa \\
\email{rendani.mbuvha@wits.ac.za, \{ilyesb,tmarwala,\}@uj.ac.za} }

\maketitle              
\begin{abstract}
Credit risk modelling is an integral part of the global financial system. While there has been great attention paid to neural network models for credit default prediction, such models often lack the required interpretation mechanisms and measures of the uncertainty around their predictions. This work develops and compares Bayesian Neural Networks(BNNs) for credit card default modelling. This includes a BNNs trained by Gaussian approximation and the first implementation of BNNs trained by Hybrid Monte Carlo(HMC) in credit risk modelling. The results on the Taiwan Credit Dataset show that BNNs with Automatic Relevance Determination(ARD) outperform normal BNNs without ARD. The results also show that BNNs trained by Gaussian approximation display similar predictive performance to those trained by the HMC. The results further show that BNN with ARD can be used to draw inferences about the relative importance of different features thus critically aiding decision makers in explaining model output to consumers. The robustness of this result is reinforced by high levels of congruence between the features identified as important using the two different approaches for training BNNs.

\end{abstract}

\section{Introduction}
 Credit default risk modelling is critical to the loss management of credit portfolios for financial institutions. Effective credit risk estimation prevents and manages losses that arise when borrowers cannot make the necessary credit repayments on agreed terms.
Accurate estimation of an individual's credit risk is of benefit both to the lending institution and the borrower in any credit agreement \cite{hand_henly}. The lending institution benefits from increased profits or reduced loss while the borrower benefits through only being involved in transactions which are within their ability of fulfilment. Subjective expert judgement has historically been used to determine the credit risk presented by a borrower\cite{Wang:2011:CAE:1860128.1860220}. This clearly presents challenges of introducing cognitive biases and does not allow for streamlined operational efficiency in large financial institutions\cite{Wang:2011:CAE:1860128.1860220}.

In recent times, increased demand for credit and the development of efficient computing systems has given prominence to sophisticated machine learning techniques in the credit risk determination process\cite{hand_henly}.
Literature in machine learning approaches to credit risk modelling is dominated by tree based models and artificial neural networks(ANNs) \cite{baesens,yeh2009comparisons}.

\citet{XIA2017225} proposed an Xtreme Gradient boosting (XGboost) tree model for credit scorecard creation. Their results show that XGboost after Bayesian parameter tuning outperforms random forests(RF) and support vector machines(SVMs) based on accuracy and the Area Under the Curve(AUC) measures on five benchmark credit datasets. \citet{twala2010multiple} shows that ensemble tree based classifiers demonstrate superior predictive performance when noise is introduced to credit scoring attributes.

\citet{sun2018} compare deep neural networks(DNNs) and simple ANNs in predicting credit default on a Brazilian Banking Dataset. Their results show that simple ANNs and DNNs outperform logistic regression and tree based methods on the AUC performance measure. 
The work of \citet{ANGELINI2008733} similarly finds that low classification errors can be obtained when using ANNs to predict credit default for Italian Small businesses. \citet{yeh2009comparisons} use ANNs to predict credit card default for Taiwanese credit cardholders and their results show that ANNs produced the lowest errors when predicting the probability of default as compared to K-nearest neighbours, logistic regression and decision trees. The work of \citet{hamori2018ensemble} shows that the performance of ANNs in credit default modelling is significantly affected by the choice of activation function and the dropout mechanism employed. 

While ANN methods are well covered in the literature, there has been no attempt to address some of their short comings when it comes to their applications to credit risk modelling. These shortcomings include the fact that traditional ANNs do not give an indication of which attributes are relevant for the prediction of credit risk and this does not aid in the transparency of the credit granting process which might be required by regulations such as General Data Protection Regulation(GDPR)'s 'right to explanation' in the European Union \cite{goodman2016eu} . The second drawback which applies to both ANNs and tree based models is that they give the probability of default but do not address the level of uncertainty behind such a probability - this does not aid in the reliability analysis and risk appetite assessment of the lending institution. In this work, we develop probabilistic formulations of ANNs through the Gaussian approximation \cite{8191129,lagazio2006} and Hybrid Monte Carlo \cite{neal2012bayesian} that will address the two shortcomings identified above.

The Bayesian formulation of ANNs using a Gaussian approximation to the posterior was first proposed by \citet{Mackay1991APB}. This method has shown superior performance in numerous applications of ANNs including in conflict analysis \cite{lagazio2006}, energy consumption modelling\cite{MacKay95probablenetworks} and wind power forecasting \citep{8191129}. 
Hybrid Monte Carlo (HMC) was first applied as a technique for sampling from Bayesian Neural Network models (BNNs) in the seminal work of \citet{neal1993bayesian}. Since then, HMC has proved to be the most effective way to obtain such samples from the exact posterior distribution of BNNs, outperforming other approximate inference methods such as Gaussian approximation of \citet{MacKay95probablenetworks} and variational inference of \citet{hinton1993keeping} (\cite{wang2013adaptive},\cite{betancourt2017conceptual}). 

BNNs have numerous advantages over traditional ANNs that are useful for decision makers in credit risk modelling. These include the ability to produce a predictive distribution rather than a point estimate of the probability to default. The ability to perform Automatic Relevance Determination(ARD) in BNNs also allows decision makers to infer the relative influences of different attributes on the network outputs - this aids in addressing the notion that Neural ANNs are "black-box" models that offer no interpretation. Finally, network regularisation arises naturally in BNNs through the prior distribution which creates a bias towards simpler regularised networks and increases network generalisation in accordance with Occam's Razor\cite{MacKay95probablenetworks}. The ideas on BNNs described above have thus far not been developed and applied in credit risk modelling.

This work aims to develop and compare BNNs trained using Gaussian approximation and HMC when applied to credit default modelling.

The rest of the paper is arranged as follows: section \ref{Mlp} describes the basic Multilayer Perceptron(MLP) neural network. Section \ref{bnn} describes the Bayesian formulation of the MLP with sections \ref{Gauss} and \ref{hmc} describing how to perform inference in BNNs using the Gaussian approximation and HMC. Section \ref{expr} sets out our experiments. Section \ref{resultss} provides and discusses the results of the experiments. Section \ref{conc} concludes and gives possible future work.

\section{The Multilayer Perceptron (MLP)} \label{Mlp}
Neural networks (NNs) are learning machines that infer input output relationships of functions with arbitrary complexity \cite{DBLP:journals/corr/EricsonM17}. Feed-forward multilayer perceptron (MLP) are a type of NNs that propagate inputs through 'hidden' layers of non-linear activation functions to generate outputs.
The output of the MLP with a single hidden layer and a single output is defined by the following composite function \cite{8191129}:
\begin{align}
     h_j(x) &= \Psi\bigg(Z_j + \sum_{i}w_{ij} x_i\bigg) \\
    o_k(x) &= b_k + \sum_{j}v_{jk} h_j(x) 
\end{align}
Where $w_{ij}$ is the weight connection for the $i^{th}$ input to the $j^{th}$ hidden unit and $v_{jk}$ is the weight connection between the $j^{th}$ hidden unit to the $k^{th}$ output. $\Psi$ is a non-linear activation function.
 
Locally Optimal weight parameters of the MLP are found by back-propagation of errors between network outputs and the ground truth realisations of the system being modelled \cite{rumelhart1986learning}. This results in maximum likelihood estimates (MLE) and Maximum A Posteriori (MAP) estimates where regularisation is included.

\section{Probabilistic Multilayer Perceptron}\label{bnn}
MLP networks can be formulated within the Bayesian Probabilistic framework such that the error function of the network can be viewed as the kernel of the posterior distribution of network parameters \cite{8191129}. In the Bayesian sense the posterior distribution of the network weight parameters given the observed data and prior distribution of weights can be derived using Bayes formula as \cite{MacKay:1992:PBF:148147.148165}:
\begin{equation}
      P(\mathbf{w}|D,\mathcal{H}) = \frac{P(D|\mathbf{w},\mathcal{H})P(\mathbf{w}|\mathcal{H})}{P(D|\mathcal{H})} 
      \label{bayes}
\end{equation}
where $P(\mathbf{w}|D,\mathcal{H})$ is the posterior probability of the weights given the data and model architecture $\mathcal{H}$. $P(D|w,H)$ is the likelihood of the data given the model. $P(w|H)$ is the prior probability of the weights. $P(D|H)$ is a marginalisation across all parameters known as the evidence. 

The posterior distribution for the MLP weights in equation \ref{bayes} becomes \cite{neal1993bayesian}:
\begin{align}
    P(\mathbf{w}|\alpha,\beta,\mathcal{H})& = \frac{1}{Z(\alpha,\beta)}\exp\big(-\alpha\big( E_W + \beta E_D\big)\big)\\
              & = \frac{1}{Z_M(\alpha,\beta)} \exp\big(-M(\mathbf{w})\big) 
\label{eq:pos}              
\end{align}
Where $\alpha E_W$ is the kernel of the prior distribution on the weights and $\beta E_D$ is the kernel of the data likelihood. 

The distribution functions above are not computationally tractable in closed form and thus we rely on approximate inference techniques such as Gaussian approximation \cite{MacKay95probablenetworks} and Hybrid Monte Carlo\cite{neal2011mcmc}.

\section{Gaussian Approximation}\label{Gauss}
 
\citet{Mackay1991APB} proposed a Gaussian Approximation to the posterior distribution in equation \ref{eq:pos} based on a second order Taylor expansion of the posterior around  $\mathbf{w_{MP}}$ as follows \cite{MacKay95probablenetworks}:
\begin{equation}
    P(\mathbf{w}|\alpha,\beta,H,D) \approx \frac{1}{Z'_M(\alpha,\beta)} \exp\bigg(-M(\mathbf{w_{MP}})-\frac{1}{2}(\mathbf{w} -\mathbf{w_{MP}})^T\mathbf{A}(\mathbf{w} -\mathbf{w_{MP}})\bigg) 
\label{eq:post}    
\end{equation}

The matrix $\mathbf{A}$ contains the second derivatives of the negative log posterior.

Assuming the Gaussian approximation holds the values of the hyper-parameters $\alpha$ and $\beta$ can be approximated by alternating between gradient decent for the weights and maximising the evidence $P(D|H)$ in equation \ref{eq:pos} which becomes tractable through the Gaussian approximation. Under these simplifying assumptions the posterior weight distribution becomes \cite{MacKay95probablenetworks}:
\begin{equation}
    P(\mathbf{w}|D,\alpha,\beta) \approx \mathcal{N}(\mathbf{w_{MP}},\mathbf{A}^{-1})
\end{equation}

Due to the central limit theorem the Gaussian Approximation is expected to be a close approximation to the true posterior as the sample size becomes large \cite{lagazio2006}.

\section{Hybrid Monte Carlo}\label{hmc}
HMC is a Markov Chain Monte Carlo technique that generates samples from the exact posterior distribution by simulating a Markov Chain with a stationary distribution, which is the the target posterior distribution of the NN. 

Traditional MCMC algorithms such as Metropolis Hastings (MH) suffer from excessive random walk behaviour - where the next state of the Markov Chain is randomly proposed from a proposal distribution. This results in inefficient sampling with low acceptance rates and low effective sample sizes.

\citet{neal1993bayesian} proposed HMC which suppresses random walk behaviour by augmenting the parameter space with auxiliary momentum variables. HMC uses the gradient information of the neural network to create a vector field around the current state giving it a trajectory towards a high probability next state. The dynamical system formed by the model parameters $w$ and the auxiliary momentum  variables $p$ is represented by the Hamiltonian $H(w,q)$ written as follows \cite{neal1993bayesian}:
\begin{equation}
    H(w,q)=M(w)+K(p)
\end{equation}
Where $M(w)$ is the negative log likelihood of the posterior distribution in equation \ref{eq:pos}, also referred to as the potential energy. $K(p)$ is the kinetic energy defined by the kernel of a Gaussian with a co-variance matrix $M$ \cite{neal2012bayesian}:
\begin{equation}
    K(p) =\frac{p^TM^{-1}p}{2}
\end{equation}.

The trajectory vector field is defined by considering the parameter space as a physical system that follows Hamiltonian Dynamics. The dynamical equations governing the trajectory of the chain are then defined by the Hamilton's equations at a fictitious time $t$ as follows \cite{neal1993bayesian}:
\begin{equation}
    \frac{\partial w_i}{\partial t} =  \frac{\partial H}{\partial p_i}
\end{equation}
\begin{equation}
    \frac{\partial p_i}{\partial t} =  -\frac{\partial H}{\partial w_i}
\end{equation}

In practical terms the dynamical trajectory is discretised using the leapfrog integrator. In the leapfrog integrator to reach the next point in the trajectory we take half a step in the momentum direction, followed by a full step in the direction of the network parameters - then ending with another other half step in the momentum direction.  
 


Finally due to the discretising errors that arise from the leapfrog integrator a Metropolis acceptance step is performed in order to accept or reject the new sample proposed by the trajectory\cite{neal2011mcmc}. In the Metropolis step the parameters proposed by the HMC trajectory $w*$ are accepted with the probability \cite{neal1993bayesian}:
\begin{equation}
        P(accept) = min\bigg(1,\frac{P(w^{*}|D,\alpha,\beta,H)}{P(w|D,\alpha,\beta,H)}\bigg)
    \end{equation}

Algorithm \ref{alg:BNN} shows the pseudo-code for the HMC where $\epsilon$ is a discretisation step-size. The leapfrog steps is repeated until the maximum trajectory length $L$.

In HMC the hyperparameters $\alpha$ and $\beta$  can be sampled by alternating between HMC samples of weights and re-sampling the regularisation parameters from an assumed distribution given a fixed sample of weights. 

\begin{algorithm}
\SetAlgoLined
\DontPrintSemicolon
\KwData{Training dataset \{$\mathbf{X}^{(i)},\mathbf{t}^{(i)}\}$}
\KwResult{$N$ Samples of MLP Weights $\mathbf{w}_{MP}$}  
\For{$n\leftarrow 1$ \KwTo $N$}{
$w_0\leftarrow w_{\text{init}}$\;
\emph{sample the auxiliary momentum variables $p$}\;
p$\sim\mathcal{N}(0,\mathbf{M})$\;
Use leapfrog steps to generate proposals for $w$\;
\For{$t\leftarrow 1$ \KwTo $L$}{
$p(t+\epsilon/2) \leftarrow p(t) + (\epsilon/2)\frac{\partial H}{\partial w}\bigg(w(t)\bigg)$\;
$w(t+\epsilon) \leftarrow w(t) + \epsilon\frac{p(t+\epsilon/2)}{M}$\;
$ p(t+\epsilon) \leftarrow p(t)+\epsilon/2) + (\epsilon/2)\frac{\partial H}{\partial w}\bigg(w(t+\epsilon)\bigg)$\;
}
\emph{Metropolis Update step:}\;
$(p,w)_n\leftarrow(p(L),w(L))$ with probability:\;
$min\bigg(1,\frac{P(w^{*}|D,\alpha,\beta,H)}{P(w|D,\alpha,\beta,H)}\bigg)$\;

}
 \caption{ Hamiltonian Monte Carlo Algorithm}
 \label{alg:BNN}
\end{algorithm}

\section{Automatic Relevance Determination(ARD)}\label{ARD}
The MLP formulations presented above can be parameterised such that different groups of weights can come from unique prior distributions and thus have unique regularisation parameters $\alpha_c$ for each class. An Automatic Relevance Determination (ARD) MLP is one where weights associated with each network input belong to a distinct class. The loss function in ARD is a follows\cite{MacKay95probablenetworks}:

\begin{equation} 
P(w|D,\alpha,\beta,H)=\frac{1}{Z(\mathbf{\alpha},\beta)}\exp\bigg(\beta E_D+\sum_{c}{\alpha_c} E_{W_C}\bigg)
\end{equation}

The regularisation hyperparameters for each class of weights can be estimated online during the inference stage of each method as discussed in sections \ref{Gauss} and \ref{hmc} . The resulting regularisation parameter $\alpha_c$ for each input can be inferred as denoting the relevance of each input. Irrelevant inputs will have high values of the regularisation parameter meaning that their weights will be forced to decay to values close to zero.

ARD thus embodies the principle of Occam's Razor by placing preference on simpler models that regularise noisy irrelevant inputs \cite{MacKay95probablenetworks}.   

\section{Experiment Setup}\label{expr}

We use the Taiwan Credit Card Dataset of \citet{yeh2009comparisons} to illustrate the utility of the various Bayesian inference methods described above. A single hidden layer MLP with 5 hidden neurons based on an initial hidden unit grid search. The following MLP inference methods are compared: A Bayesian MLP trained by HMC, a Bayesian MLP with ARD trained by HMC and a Bayesian MLP with ARD trained by Gaussian Approximation.

\subsection{Dataset}

The dataset consists of 30000 records with 23 features relating to the demographic profile of credit cardholders as well as their historical payment behaviour. The target variable is a binary indicator of whether the client has had a default payment. The dataset consisted of the following features:  
\begin{table}
\centering 
\caption{Attributes in the Taiwanese Credit Card Dataset}
\begin{tabular}{|c |c|} 
\hline 

\textbf{Attribute } & \textbf{Attribute Name}  \\
\hline
X1 & Amount of the given credit\\
X2 & Gender \\
X3 & Education \\ 
X4 & Marital status\\  
X5 & Age (years)\\ 
X6 - X11 & History of past payment - each of the last six months \\
X12 - X17 & Amount of bill statement - each of the last six months\\ 
X18 - X23 & Amount of previous payment - each of the last six months \\
\hline 
\end{tabular}
 
\label{table:ft1} 
\end{table}

Normalisation of attributes has been shown to improve the performance of MLP models\cite{hamori2018ensemble}. Thus, each of the attributes above is  pre-processed by projecting it onto the range $[0,1]$ using min-max normalisation.

A random split 70:30 training and testing partition of the data is used for all the models.

\subsection{Performance Evaluation}
The Area Under the Receiver Operating Characteristic Curve (AUC) and a confusion matrix are used to evaluate predictive performance.

The AUC can be interpreted as the probability of the model correctly assigning higher probability of default to defaulters relative to non-defaulters. The AUC has been shown to be a more robust metric in unbalanced classification problems where datasets are often biased towards non-defaulters \cite{ANGELINI2008733}.

.

\section{Results}\label{resultss}
\subsection{Predictive Performance}
\begin{figure}[h]
    \centering
    \includegraphics[width=1\textwidth]{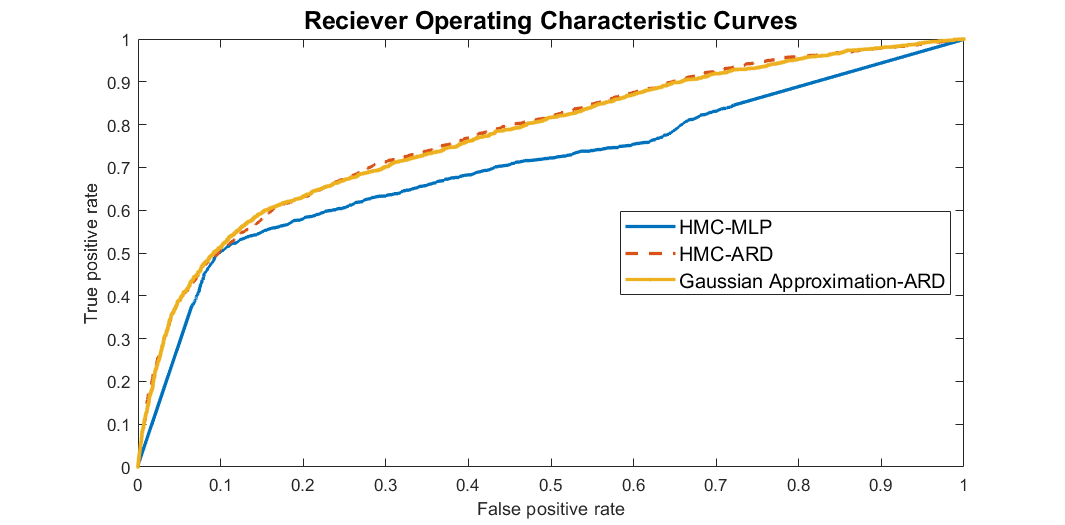}
    \caption{Receiver Operating Characteristics Curves of the various models on the Taiwan Credit Card Dataset}
    \label{fig:roc_curve}
\end{figure}

We now discuss the performance results on the testing set of the models specified in section \ref{expr}. Figure \ref{fig:roc_curve} shows the receiver operating characteristic(ROC) curves of each of the classifiers. The ROC curve shows that the performance of the HMC-ARD with an AUC of 0.7783 and Gaussian approximation-ARD with an AUC of 0.7753 both outperform the HMC-MLP with an AUC of 0.7079. This result implies that adding the ARD to the BNN results in a more accurate model. This effect could be attributed to the fact that the ARD regularises noisy inputs thus leaving the network with greater generalisation ability. It is important to note that while the AUC gives equal weight to miss-classifications of defaulters and non-defaulters, this is not necessarily the case in credit risk modelling where the cost of miss-classifying a default (false negative) is much higher than that of miss-classifying a non-default.   

Table \ref{table:Results1} gives some insight into the miss-classification rates of the different models. As can be seen from the table the HMC-ARD model seems to marginally have a lower number of false negatives than the Gaussian approximation ARD model. This would save the lending institution from advancing credit to a borrower who is unlikely to repay the loan. While the differences in predictive performance is marginal, it is important to highlight that the HMC-ARD is more theoretically robust than the Gaussian approximation as it samples the exact posterior distribution. 

\begin{table}[!h]
\centering 
\caption{Table showing performance results on the testing set for Taiwan Credit Card Dataset}
\begin{tabular}{|c |c |c|c|c|} 
\hline
Model  & True Default  &  False Non-Defaults & True Non-Defaults & False Defaults \\
\hline

HMC-MLP & 934 & 614 & 6407 & 1045 \\
Gaussian Approximation-ARD & 772 & 355 & 6666 & 1207 \\
HMC-ARD  & 754 & 331 & 6690 & 1225 \\
 \hline 
\end{tabular}
\label{table:Results1} 
\end{table}
\subsection{ARD Output and Model Interpretation} 
 
\begin{figure}
    \centering
    \includegraphics[width=0.98\textwidth]{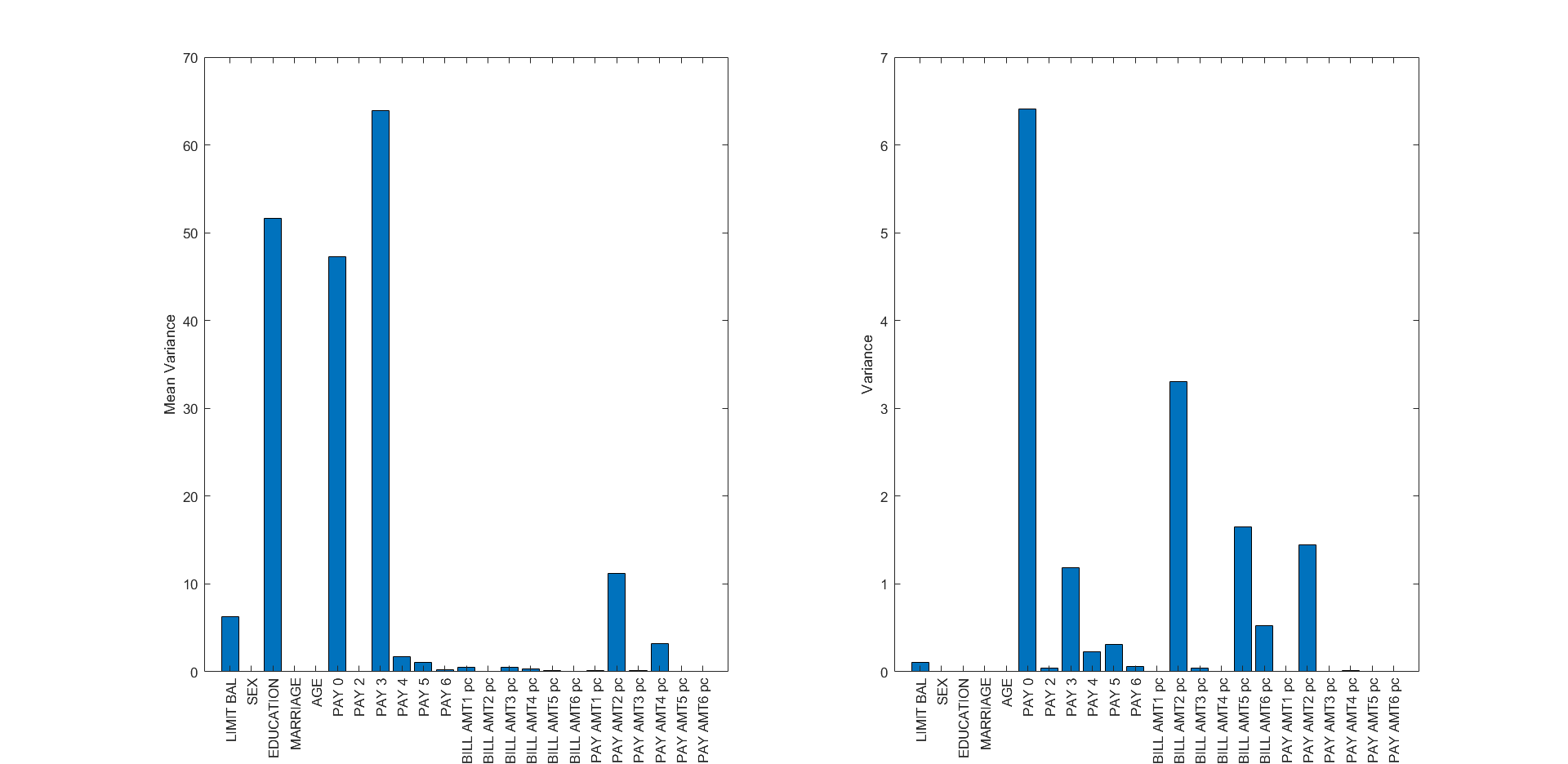}
    \caption{Left, mean posterior variances from the HMC-ARD model which indicate the relevance of each attribute. Right, posterior variances from the Gaussian Approximation-ARD model which indicate the relevance of each attribute}
    \label{fig:ard_compare}
\end{figure}

\begin{figure}
    \centering
    \includegraphics[width=0.98\textwidth]{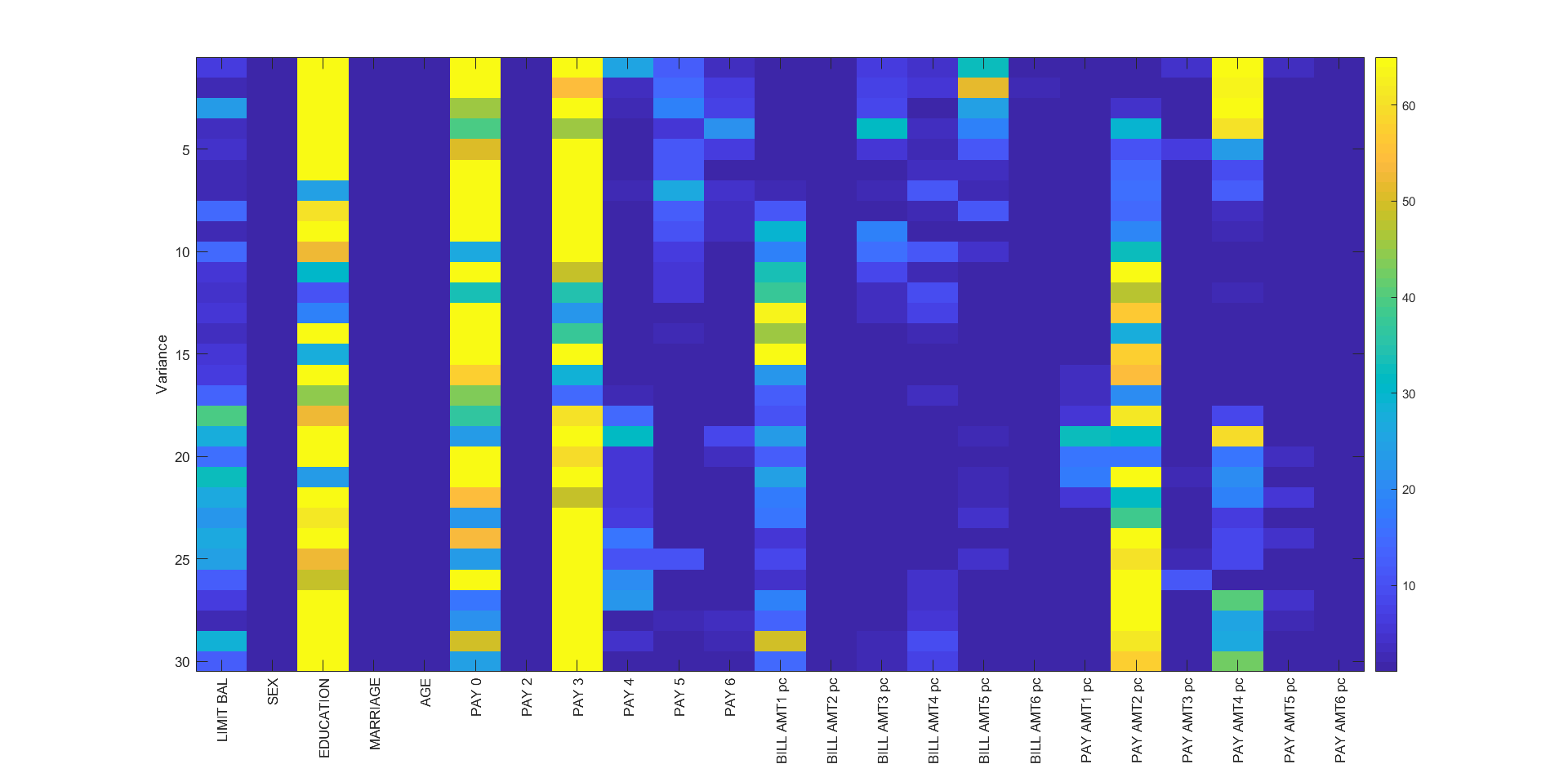}
    \caption{A heat-map showing the exact samples of posterior variances from the HMC-ARD model}
    \label{fig:heat}
\end{figure}

We now explore the output of the ARD to draw inferences on the relative influences of different features on the output. This feature of BNNs is critical to credit risk modelling as it allows lending institutions to understand the drivers of credit risk and, therefore, puts them in a position that affords their clients the 'right to explanation'. Figure \ref{fig:ard_compare} shows a comparison of the ARD output of HMC-ARD and the Gaussian approximation. Since the HMC-ARD model gives samples of the posterior variances we use the mean posterior variances for the comparisons, the heap-map of the exact samples from the posterior can be seen in figure \ref{fig:heat}. Both models identify the payment status six months ago (PAY 0) as the most important factor. The models also seem to agree on other features that are identified as important such as payment status 3 months ago (PAY 3), amount of credit given(LIMIT BAL) and the amount of previous payment 4 months ago (PAY AMT 2). The HMC-ARD model also highlights the education level as a relevant predictor.

\section{Conclusion}\label{conc}
We developed and compared two approaches for Bayesian inference in neural networks. The results show that the Gaussian approximation-ARD and the HMC-ARD models display similar predictive performance with clear out-performance over the basic HMC-MLP. The more theoretically grounded HMC-ARD does show a marginally superior ability to manage false negatives which is ideal for credit data.

Both models crucially also allow for interpretation of the relative feature influences on the probability of default. The ARD results of both models show significant overlap which indicates some robustness in the relevance determination ability of the Bayesian frameworks \cite{lagazio2006}.  

Future improvements to this work could include exploring alternative variants of the HMC sampling such as the Riemannian Manifold Hamiltonian Monte Carlo of \citet{girolami2011riemann}. Deeper neural networks could also be explored.

%

%
%
%
\bibliographystyle{plainnat}
\bibliography{references}

\end{document}